\documentclass[runningheads]{llncs}
\usepackage[T1]{fontenc}
\usepackage{graphicx}
\usepackage{amssymb}
\usepackage{multirow}
\usepackage{booktabs}
\usepackage{pgfplots}
\usepgfplotslibrary{groupplots} 
\pgfplotsset{compat=1.18}
\usepackage{microtype}
\usepackage[hyphens]{url}
\usepackage[pagebackref=true,breaklinks=true]{hyperref}
\hypersetup{
     colorlinks,
     linkcolor={red!75!black},
     citecolor={blue!75!black},
     urlcolor={blue!75!black},
}
\usepackage{color}

\usepackage[nolist]{acronym}

\begin{acronym}
    \acro{BC}{In-Batch Contrastive}
    \acro{CER}{Character Error Rate}
    \acro{CNN}{Convolutional Neural Network}
    \acro{CTC}{Connectionist Temporal Classification}
    \acro{EC}{Error-Based Contrastive}
    \acro{ECHWR}{Error-Aware Contrastive-Enhanced Handwriting Recognition}
    \acro{IMU}{Inertial Measurement Unit}
    \acro{LSTM}{Long Short-Term Memory}
    \acro{MACs}{Multiply-Accumulate Operations}
    \acro{OnHWR}{Online Handwriting Recognition}
    \acro{RMS}{Root Mean Square}
    \acro{WD}{Writer-Dependent}
    \acro{WER}{Word Error Rate}
    \acro{WI}{Writer-Independent}
\end{acronym}
\begin{document}
\title{Enhancing IMU-Based Online Handwriting Recognition via Contrastive Learning with Zero Inference Overhead}
\titlerunning{Enhancing IMU-Based OnHWR via CL with Zero Inference Overhead}
%
\author{Jindong Li\inst{1, 4}\orcidID{0000-0002-3550-1660} \and
    Dario Zanca\inst{1}\orcidID{0000-0001-5886-0597} \and
    Vincent Christlein\inst{1}\orcidID{0000-0003-0455-3799} \and
    Tim Hamann\inst{2}\orcidID{0000-0003-3562-6882} \and
    Jens Barth\inst{2}\orcidID{0000-0003-3967-9578} \and
    Peter Kämpf\inst{2} \and
    Björn Eskofier\inst{1, 3, 4, 5}\orcidID{0000-0002-0417-0336}}

\authorrunning{J. Li et al.}
%
\institute{Friedrich-Alexander-Universität Erlangen-Nürnberg, Erlangen, Germany \and STABILO International GmbH, Heroldsberg, Germany \and Ludwig-Maximilians-Universität München, Munich, Germany \and Munich Center for Machine Learning, Munich, Germany \and Helmholtz Zentrum München - German Research Center for Environmental Health, Neuherberg, Germany}

\maketitle              
\begin{abstract}

    Online handwriting recognition from inertial measurement units enables handwriting on paper as input for digital devices. Running recognition on edge hardware improves privacy and lowers latency, but entails memory constraints. To address this, we propose Error-Aware Contrastive-Enhanced Handwriting Recognition (ECHWR), a training-time contrastive learning framework designed to improve recognition accuracy without increasing inference costs. During training, ECHWR utilizes a temporary auxiliary branch that aligns IMU signal embeddings with semantic text embeddings in a shared space. This is enforced with a dual objective: (i) an in-batch contrastive loss that aligns IMU and text embeddings, and (ii) a hard-negative contrastive loss that uses synthetically perturbed text targets. The auxiliary branch is discarded after training, which allows the deployed model to keep its original, efficient architecture. Experiments on the OnHW-Words500 dataset show that ECHWR significantly outperforms state-of-the-art baselines, reducing character error rates by up to 7.4\% on the writer-independent split and 10.4\% on the writer-dependent split. Ablations indicate that the hard-negative objective is particularly effective for generalization to unseen writing styles, making our approach well-suited for online handwriting recognition on unseen users. Code is available at: \url{https://github.com/jindongli24/ECHWR}.

\end{abstract}

\begin{figure}[ht]
    \centering
    \includegraphics[width=0.95\linewidth]{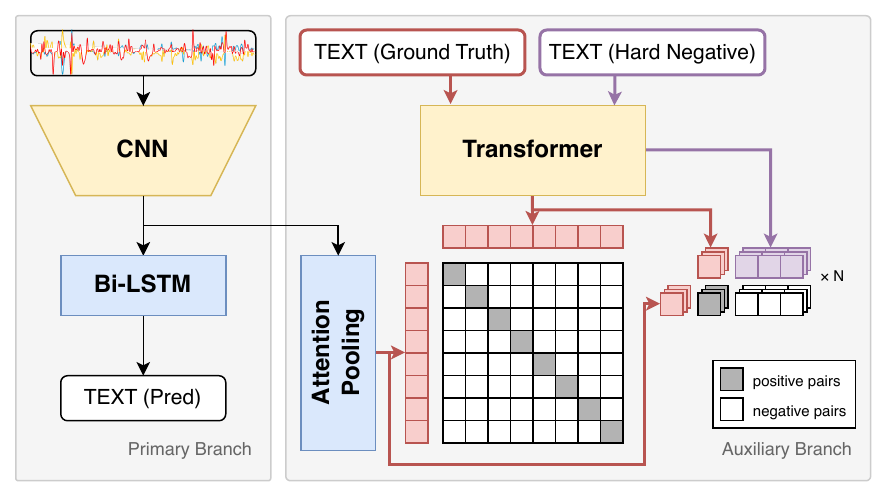}
    \caption{\textbf{The \ac{ECHWR} framework.} The model trains a primary branch (left) jointly with an auxiliary branch to align feature representations. The training objective combines \acs{CTC} loss with two contrastive components: an in-batch contrastive loss (middle) to align matching sensor-text pairs and a novel error-based contrastive loss (right) to discriminate against synthetic ``hard negatives.'' The auxiliary branch is removed during inference to maintain zero computational overhead.}
    \label{fig:framework}
\end{figure}

\section{Introduction}

Handwriting remains a highly natural and efficient form of human communication. This makes \ac{OnHWR}, particularly \ac{IMU}-based systems, a critical interface for modern smart devices. Because dynamic handwriting data contains unique biometric traits that can reveal sensitive user information~\cite{privacy1,privacy2}, users are often hesitant to transmit raw sensor trajectories to cloud servers. However, executing recognition locally imposes strict computational limits. While large-scale models offer high accuracy, they are frequently unsuitable for edge devices due to latency and memory constraints. Consequently, there is a critical need for efficient \ac{OnHWR} approaches that maintain high performance while staying within the hardware capabilities of portable electronics.

Meeting the privacy, latency, and memory requirements of on-device recognition limits the use of large or computationally heavy models. Although modern encoder-decoder architectures can achieve strong accuracy, many rely on complex components that are costly at inference time~\cite{origaminet,trocr,htr-vt}. In resource-constrained environments, simply increasing model size to improve feature extraction is therefore impractical. This motivates training strategies that improve recognition accuracy while keeping the deployed inference cost unchanged.

Contrastive learning has become a standard technique in computer vision for learning robust features by aligning similar representations~\cite{simclr,clip}, and this success has extended to time-series analysis~\cite{cl_pt_1}. However, these methods typically use contrastive learning only for pre-training, which is also true for sequence recognition tasks such as speech or \ac{IMU}-based handwriting recognition. Consequently, the potential for contrastive learning to enhance feature quality directly during the main training phase remains largely unexplored.

To bridge this gap, we propose \ac{ECHWR}, a training-time contrastive learning framework designed to improve \ac{OnHWR}  accuracy without increasing inference costs. Unlike traditional approaches trained solely with \acf{CTC} loss, we augment the model during training with a temporary auxiliary projection branch that maps the IMU signal features into a shared embedding space with semantic text embeddings produced by a Transformer text encoder from the ground-truth transcript. We enforce sensor-text alignment using a dual contrastive objective. (i) An in-batch contrastive loss pulls together matched sensor-text pairs and pushes apart mismatched pairs within the mini-batch. (ii) A ``hard negatives'' contrastive loss further improves discrimination by separating the correct transcript from synthetically perturbed alternatives created via random deletions, insertions, and substitutions. Figure~\ref{fig:framework} offers a comprehensive overview of the method. Crucially, we discard the auxiliary branch after training. This approach allows the primary \ac{IMU}-based \ac{OnHWR} model to learn richer features and achieve higher accuracy without increasing computational costs or altering the architecture during deployment.

\section{Related Works}
\subsection{\ac{IMU}-Based \ac{OnHWR}}

Early \ac{IMU}-based handwriting recognition relied heavily on manual feature engineering and statistical modeling. Researchers typically employed sensor fusion techniques, such as quaternion-based complementary filters, to reconstruct motion trajectories from raw accelerometer and gyroscope signals~\cite{imu_dtw_1}. Dynamic time warping served as the standard approach for handling temporal variations in isolated characters~\cite{imu_dtw_1,imu_dtw_2}. For more complex sequences, probabilistic models like hidden Markov models were used for both character classification~\cite{imu_hmm_dtw} and continuous text spotting in 3D space~\cite{imu_svm_hmm}. Despite their utility, these methods often struggled with sensor noise, sensor bias, and the significant variability found in individual writing styles.

The advent of deep learning shifted the focus toward end-to-end neural networks that extract features directly from raw data. Architectures combining \acp{CNN} and \ac{LSTM} networks became the standard for capturing local and temporal dependencies. By using the \ac{CTC} loss, these models align unsegmented sensor streams with text, enabling robust word and sentence recognition~\cite{imu_cnn_lstm_1,imu_cnn_lstm_2}. More recently, research has shifted toward efficiency, driven by the need to deploy recognition on resource-constrained devices. This includes exploring lightweight architectures and specialized training strategies designed to maintain high performance without the heavy computational costs typically associated with deep models~\cite{rewi}.

\subsection{Contrastive Learning}

Modern representation learning relies heavily on instance discrimination tasks. Methods such as SimCLR~\cite{simclr} and CLIP~\cite{clip} use the InfoNCE loss to maximize agreement between different versions of the same image while increasing the distance from other samples in the batch. By optimizing this objective, these frameworks encourage the model to learn features that remain consistent across transformations but are still distinct enough to discriminate between different instances.

In the time-series domain, researchers have adapted these objectives to capture temporal dynamics. For example, TS-TCC~\cite{cl_pt_1} and other pre-training frameworks~\cite{cl_pt_2,cl_pt_3,cl_pt_4} typically enforce consistency between different augmentations or time-frequency views of a signal. These methods generally treat contrastive learning as a standalone pre-training phase used to initialize encoder weights. In contrast, joint optimization strategies are still rare. While a few studies in spatio-temporal forecasting~\cite{cl_mt_1} and human activity recognition~\cite{cl_mt_2} have used contrastive loss as an auxiliary regularizer, these works focus on classification rather than sequence-to-sequence recognition with structured decoding objectives (e.g., \ac{CTC} loss).

Regardless of the domain, the effectiveness of contrastive objectives depends heavily on the sampling strategy. Traditional approaches that use random ``in-batch'' negatives often yield a weak training signal once most negatives become trivially separable, as the model can easily distinguish the majority of negative pairs. Research in hard negative mining~\cite{hard_neg_1,hard_neg_2,hard_neg_3} shows that prioritizing difficult samples prevents the model from learning simple or trivial solutions. By focusing on samples that are semantically different yet close in the embedding space, these methods force the encoder to learn more detailed and discriminative features.

\section{Methods}
\subsection{Model Architecture}

In our experiments, we adopt REWI~\cite{rewi} as the underlying recognition backbone, motivated by its state-of-the-art accuracy and efficiency for \ac{IMU}-based \ac{OnHWR}. This model follows an encoder-decoder architecture trained with \ac{CTC} loss, where the encoder is a \ac{CNN} and the decoder is a network based on bidirectional \ac{LSTM}.

To improve the feature representation of the \ac{CNN} backbone without changing its structure for inference, we add a temporary auxiliary branch. This branch aligns sensor signals with semantic text embeddings and consists of a dedicated text encoder and an attention pooling module to bridge the difference in dimensionality between the time-series sensor features and the text representations. During training, the model is regularized by two auxiliary objectives: an in-batch contrastive loss to align sensor and text representations, and an error-based contrastive loss to help the model distinguish against hard artificial negative samples. Importantly, the entire auxiliary branch is discarded after training. This ensures that the deployed model's architecture remains identical to the original REWI backbone and operates with no additional computational costs.

\subsubsection{Attention Pooling}

The attention pooling layer maps the variable-length sensor sequence to a fixed-length representation that can be aligned with the corresponding text embedding. First, the primary \ac{CNN} encoder produces a temporal sequence of features $X \in \mathbb{R}^{T \times D_{in}}$, where $T$ represents the sequence length. The module then linearly projects these features to a target dimension $D_{out}$ of 512. Standard sinusoidal positional embeddings are added at this stage to preserve the temporal order of the data.

Finally, a multi-head attention mechanism with 8 heads is applied. In this step, the mean of the projected sequence serves as the \textit{query} ($Q$), while the entire sequence serves as both the \textit{keys} ($K$) and \textit{values} ($V$). This process aggregates relevant sensor segments into a single global token, resulting in a fixed-length context vector $c_{sig} \in \mathbb{R}^{D_{out}}$ for contrastive alignment.

\subsubsection{Text Encoder}

The text encoder generates semantic embeddings from both ground-truth and hard negative transcripts. We employ a lightweight Transformer with three encoder layers~\cite{trans}, configured with 8 attention heads, a model and embedding dimension of 512, and an attention dropout rate of 0.1. The encoder operates at the character level, treating each character as an individual token to match the fine-grained nature of handwriting. To capture spatial relationships, learnable position embeddings are added to the token embeddings. We also prepend a learnable \texttt{[CLS]} token to the input sequence to aggregate global context. The final output vector of this token serves as the global text embedding $z_{text} \in \mathbb{R}^{D_{out}}$. This encoder is trained from scratch in conjunction with the sensor network.

\subsubsection{Representation refinement}

As embedding quality is essential for contrastive learning, we investigate the effect of the gated attention mechanism~\cite{gated}, registers~\cite{register}, and \ac{RMS} normalization~\cite{rmsnorm} on feature representation quality. While the first two methods originate from large language models and computer vision, respectively, they are both designed to mitigate the issue of attention-related artifacts. The gated attention mechanism adds a gating factor to the attention formulation, allowing the model to focus on informative features while suppressing noise. Registers are auxiliary learnable tokens added to the input sequence that can act as ``sinks'' for global information artifacts, allowing global context to be learned without overloading content tokens. This prevents the model from repurposing semantic tokens for global reasoning. Additionally, we employ \ac{RMS} normalization to improve training stability and representation quality. We apply head-wise gated attention to both the attention pooling layer and the text Transformer encoder, whereas \ac{RMS} normalization and registers are only used within the text Transformer. We evaluate these design choices in Sec.~\ref{sec:architectural_analysis}.

\subsection{Training Objective}

As shown in Eq.~\ref{eq:loss}, we use a composite objective function designed to jointly optimize handwriting recognition accuracy and semantic representation alignment. It integrates a smoothed \ac{CTC} loss adopted from REWI~\cite{rewi}, with two distinct contrastive mechanisms. Specifically, we employ an \ac{BC} loss to align time-series and textual modalities, and an \ac{EC} loss to help the model distinguish ground-truth text from hard negative errors.

\begin{equation}
    \mathcal{L}_{total} = \mathcal{L}_{CTC} + \mathcal{L}_{BC} + \mathcal{L}_{EC}
    \label{eq:loss}
\end{equation}

\subsubsection{In-Batch Contrastive Loss}

The \ac{BC} loss aligns the latent representations of time-series sensor data with their corresponding textual transcripts. Given a batch of $N$ sensor-text pairs, let $c_{sig}^{(i)}$ and $z_{text}^{(i)}$ represent the normalized sensor and text embeddings for the $i$-th sample. The similarity between a sensor embedding and a text embedding is computed as $s_{i,j} = \tau \cdot (c_{sig}^{(i)\top} z_{text}^{(j)})$, where $\tau$ is a learnable scaling parameter. The loss is then calculated symmetrically as follows:

\begin{equation}
    \mathcal{L}_{BC} = - \frac{1}{2N} \sum_{i=1}^{N} (\log \frac{\exp(s_{i,i})}{\sum_{j=1}^{N} \exp(s_{i,j})} + \log \frac{\exp(s_{i,i})}{\sum_{j=1}^{N} \exp(s_{j,i})})
\end{equation}

To avoid false negatives caused by identical transcripts, we filter the batch to ensure that each unique text label appears only once.

\subsubsection{Error-Based Contrastive Loss}

The \ac{EC} loss forces the model to distinguish ground-truth text from artificial hard negatives. For each sample $i$, we generate $S$ sets of hard negatives. Each set contains three variants that correspond to single-character deletion, insertion, or substitution errors. This process results in a total of $M=3S$ negative samples $\{z_{text}^{(i,k)}\}_{k=1}^{M}$, all of which have a Levenshtein distance of one from the ground truth $z_{text}^{(i,0)}$. The loss is computed as follows:

\begin{equation}
    \mathcal{L}_{EC} = - \frac{1}{N} \sum_{i=1}^{N} \log \frac{\exp(\tau \cdot c_{sig}^{(i)\top} z_{text}^{(i,0)})}{\sum_{k=0}^{M} \exp(\tau \cdot c_{sig}^{(i)\top} z_{text}^{(i,k)})}
\end{equation}

This objective compels the model to learn fine-grained features. These features are necessary to accurately discriminate between highly similar transcripts that differ by only a single character.

\section{Experiments}
\subsection{Datasets}

Public \ac{IMU}-based \ac{OnHWR} datasets are rare, especially for word-level recognition. To our knowledge, OnHW-Words500~\cite{imu_cnn_lstm_2} is the only publicly available word-based dataset. We use its larger, right-handed subset for all our experiments. This subset contains 13-channel handwriting data collected from 53 subjects using a sensor-enhanced pen developed by STABILO International GmbH. The dataset defines two distinct evaluation protocols that use 5-fold cross-validation, namely \ac{WD} (split by words) and \ac{WI} (split by writers). We use both protocols to evaluate how our model handles different distribution shifts between the training and test sets, specifically variations in character distribution (illustrated in Appx.~\ref{app:visualize_character_distribution}) and differences in individual handwriting styles.

\subsection{Implementation Details}

We adopt the core architecture from REWI alongside its 1-D encoder variants of ConvNeXt and Swin Transformer V2~\cite{rewi}. These variants were specifically adapted for time-series data and have a similar capacity to the base REWI encoder. Following a similar setup, we train the model for 300 epochs with a batch size of 64. The training process employs a learning rate schedule consisting of a 30-epoch linear warmup followed by cosine annealing. We use the AdamW optimizer with a weight decay of $10^{-2}$. For the primary \ac{OnHWR} backbone, we set the base learning rate to $10^{-3}$, whereas we use a lower base learning rate of $2.5 \times 10^{-4}$ for the auxiliary contrastive branch to ensure training stability. The framework is implemented using PyTorch 2.9.1 on an NVIDIA RTX 3090 GPU with 24 GB of VRAM.

We evaluate performance at two levels of granularity. \ac{CER} measures errors at the character level, while \ac{WER} assesses them at the word level. These metrics quantify the total number of substitutions, deletions, and insertions required to align the model's predictions with the ground truth. In our experiments, we always present \ac{CER} and \ac{WER} as percentages. To provide a comprehensive view of computational complexity alongside recognition accuracy, we also report the total number of parameters and the \ac{MACs}.

\subsection{Loss Terms Analysis}

To rigorously assess the effectiveness of the \ac{ECHWR} objectives, we evaluated performance across primary branches with various architectures, capacity, and loss term configurations, as shown in Table~\ref{tab:objective_contribution}.

\begin{table}[ht]
    \centering
    \addtolength{\tabcolsep}{3pt}
    \caption{\textbf{Loss Terms Analysis.} The \ac{WD} split results are based on the \textit{RMSNorm + Gated Attention} architecture, while the \ac{WI} split uses the \textit{LayerNorm + Gated Attention + Registers} architecture. Suffixes "-D$*$" indicate the number of layers in each encoder block. Models with REWI/B, ConvNeXt, and Swin Transformer V2 encoders use the base version of the REWI BiLSTM decoder, while other variants use the small version. The numbers of parameters (\#Params) and \ac{MACs} are for inference time only.}
    \begin{tabular}{lccccccccc}
        \toprule
        \multirow{2}{*}{Encoders}                 & \multicolumn{2}{c}{Aux. Obj.} & \multicolumn{2}{c}{\ac{WD}} & \multicolumn{2}{c}{\ac{WI}} & \multirow{2}{*}{\#Params} & \multirow{2}{*}{MACs}                                                                   \\
                                                  & \ac{BC}                       & \ac{EC}                     & CER                         & WER                       & CER                   & WER            &                        &                       \\
        \midrule
        \multirow{3}{*}{REWI/S}                   &                               &                             & 15.99                       & 51.17                     & \textbf{10.24}        & \textbf{24.55} & \multirow{3}{*}{0.54M} & \multirow{3}{*}{80M}  \\
                                                  & \checkmark                    &                             & \textbf{15.59}              & \textbf{50.28}            & 10.61                 & 25.47          &                        &                       \\
                                                  & \checkmark                    & \checkmark                  & 15.99                       & 50.77                     & 10.59                 & 25.18          &                        &                       \\
        \cmidrule(lr){2-7}
        \multirow{3}{*}{REWI/S-D2}                &                               &                             & 15.40                       & \textbf{48.93}            & 9.38                  & 21.66          & \multirow{3}{*}{0.72M} & \multirow{3}{*}{111M} \\
                                                  & \checkmark                    &                             & 15.46                       & 49.28                     & \textbf{9.13}         & \textbf{20.65} &                        &                       \\
                                                  & \checkmark                    & \checkmark                  & \textbf{15.36}              & 49.09                     & 9.27                  & 21.17          &                        &                       \\
        \cmidrule(lr){2-7}
        \multirow{3}{*}{REWI/S-D3}                &                               &                             & 16.11                       & 50.10                     & 8.91                  & 20.12          & \multirow{3}{*}{0.91M} & \multirow{3}{*}{143M} \\
                                                  & \checkmark                    &                             & 15.73                       & 49.76                     & 8.98                  & 20.14          &                        &                       \\
                                                  & \checkmark                    & \checkmark                  & \textbf{15.62}              & \textbf{48.84}            & \textbf{8.87}         & \textbf{19.78} &                        &                       \\
        \cmidrule(lr){2-7}
        \multirow{3}{*}{REWI/B-D1}                &                               &                             & \textbf{13.58}              & \textbf{46.09}            & 10.50                 & 25.28          & \multirow{3}{*}{1.45M} & \multirow{3}{*}{225M} \\
                                                  & \checkmark                    &                             & 13.67                       & 46.51                     & \textbf{10.37}        & 25.19          &                        &                       \\
                                                  & \checkmark                    & \checkmark                  & 13.61                       & 46.19                     & 10.49                 & \textbf{25.01} &                        &                       \\
        \midrule
        \multirow{3}{*}{REWI/B}                   &                               &                             & 14.45                       & 43.96                     & 7.33                  & 15.16          & \multirow{3}{*}{3.93M} & \multirow{3}{*}{605M} \\
                                                  & \checkmark                    &                             & \textbf{12.95}              & \textbf{40.26}            & 7.03                  & 14.31          &                        &                       \\
                                                  & \checkmark                    & \checkmark                  & 14.04                       & 41.99                     & \textbf{6.79}         & \textbf{13.65} &                        &                       \\
        \cmidrule(lr){2-7}
        \multirow{3}{*}{ConvNeXt~\cite{convnext}} &                               &                             & 14.90                       & 45.99                     & \textbf{8.02}         & 17.93          & \multirow{3}{*}{3.86M} & \multirow{3}{*}{600M} \\
                                                  & \checkmark                    &                             & \textbf{11.34}              & \textbf{38.21}            & 8.04                  & \textbf{17.73} &                        &                       \\
                                                  & \checkmark                    & \checkmark                  & 12.08                       & 38.95                     & 8.18                  & 17.98          &                        &                       \\
        \cmidrule(lr){2-7}
        \multirow{3}{*}{SwinV2~\cite{swinv2}}     &                               &                             & 13.49                       & 44.33                     & 8.43                  & 19.20          & \multirow{3}{*}{3.88M} & \multirow{3}{*}{601M} \\
                                                  & \checkmark                    &                             & \textbf{11.46}              & \textbf{38.49}            & \textbf{7.99}         & 17.76          &                        &                       \\
                                                  & \checkmark                    & \checkmark                  & 11.53                       & 38.75                     & 8.02                  & \textbf{17.76} &                        &                       \\
        \bottomrule
    \end{tabular}
    \label{tab:objective_contribution}
\end{table}

The results demonstrate that the contribution of the auxiliary branch is dependent on the size, especially the depth, of the encoders. For small REWI encoders (REWI/S, 0.54M--0.91M parameters), the contribution of the auxiliary branch increases from negative to positive with the depth of the encoders. In contrast, although the shallow base REWI encoder REWI/B-D1 (1.45M parameters) is larger than others, the contribution of \ac{ECHWR} is still not as significant as the smaller but deeper encoder REWI/S-D3. Lacking sufficient parameters and depth, these smaller encoders appear saturated by the dual burden of minimizing the \ac{CTC} loss while simultaneously optimizing the auxiliary contrastive alignment.

Once the model reaches sufficient capacity, the framework delivers robust and consistent improvements. The benefits diverge depending on the specific generalization challenge. On the \ac{WD} split, using the \ac{BC} term alone allows the \ac{ECHWR} framework to reduce the \ac{CER} to 12.95\% ($-1.5$\,pp; $-10.4$\%) and the \ac{WER} to 40.26\% ($-3.7$\,pp; $-8.4$\%). This pattern of substantial improvement extends to the 1-D variants of modern vision backbones as well. Introducing the \ac{BC} term drops the ConvNeXt baseline \ac{CER} to 11.34\% ($-3.56$\,pp; $-23.9$\%) and its \ac{WER} to 38.21\% ($-7.78$\,pp; $-16.9$\%). Meanwhile, the Swin Transformer V2 encoder achieves a \ac{CER} of 11.46\% ($-2.03$\,pp; $-15.0$\%) and a \ac{WER} of 38.49\% ($-5.84$\,pp; $-13.2$\%). Notably, across all these high-capacity models, this \ac{BC}-only configuration consistently outperforms the variant that includes the \ac{EC} on the \ac{WD} split. This suggests that when writing styles are familiar but the vocabulary is novel, the primary challenge is flexibly aligning sensor signals to character representations. The \ac{EC} term appears to create a rigidity that hinders the model's ability to generalize to novel character sequences.

Conversely, the \ac{WI} split benefits from the stricter decision boundaries enforced by the \ac{EC} term. For the REWI/B encoder, while adding \ac{BC} provides a strong foundation, the full configuration with \ac{EC} yields the best overall performance, reducing the \ac{CER} to 6.79\% ($-0.54$\,pp; $-7.4$\%) and the \ac{WER} to 13.65\% ($-1.51$\,pp; $-10.0$\%). The Swin Transformer V2 encoder supports this dynamic at the word level; deploying the full configuration with the \ac{EC} term drives its baseline \ac{WER} down to an optimal 17.76\% ($-1.44$\,pp; $-7.5$\%) while reducing the \ac{CER} to 8.02\% ($-0.41$\,pp; $-4.9$\%). When facing known words written by unknown subjects, the primary source of error is often ambiguity caused by biometric variance. The \ac{EC} term forces the model to reject hard synthetic negatives, effectively cutting through the noise of unfamiliar writing styles to maximize precision on known vocabulary.

\subsection{Architectural Analysis}
\label{sec:architectural_analysis}

To better understand the mechanisms driving the observed performance gains, we analyze the impact of normalization and representation enhancement strategies. Table~\ref{tab:architectural_variations} summarizes the performance of these architectural variants across both evaluation protocols. Rather than identifying a single universally optimal configuration, the results indicate distinct architectural preferences depending on the underlying generalization challenge.

\begin{table}[ht]
    \centering
    \addtolength{\tabcolsep}{3pt}
    \caption{\textbf{Architectural Analysis of \ac{ECHWR} Variants.} We investigate the impact of various normalization strategies alongside the inclusion of head-wise gated attention (GA) and register tokens (Reg). This analysis is conducted using the REWI/B primary branch under the supervision of different \ac{ECHWR} objectives. The numbers of parameters (\#Params), \ac{MACs}, and seconds per iteration (Sec/Iter) are measured for training time only.}
    \begin{tabular}{ccccccccccc}
        \toprule
        \multicolumn{2}{c}{Variants} & \multicolumn{2}{c}{Aux. Obj.} & \multicolumn{2}{c}{\ac{WD}} & \multicolumn{2}{c}{\ac{WI}} & \multirow{2}{*}{\#Params} & \multirow{2}{*}{MACs} & \multirow{2}{*}{Sec/Iter}                                           \\
        GA                           & Reg                           & \ac{BC}                     & \ac{EC}                     & CER                       & WER                   & CER                       & WER                                     \\
        \midrule
        \multicolumn{4}{l}{REWI/B}   & 14.45                         & 43.96                       & 7.33                        & 15.16                     & 3.93M                 & 605M                      & 8.63                                    \\
        \midrule
        \multicolumn{8}{l}{\textit{LayerNorm Variants}}                                                                                                                                                                                                    \\
                                     &                               & \checkmark                  &                             & 13.66                     & 41.51                 & 7.00                      & 13.93          & 10.50M & 1.05B & 12.14 \\
                                     &                               & \checkmark                  & \checkmark                  & 13.81                     & 41.74                 & 7.10                      & 14.31          & 10.50M & 1.87B & 13.05 \\
        \checkmark                   &                               & \checkmark                  &                             & 14.51                     & 42.52                 & 6.93                      & 13.98          & 14.71M & 1.32B & 12.52 \\
        \checkmark                   &                               & \checkmark                  & \checkmark                  & 13.94                     & 41.82                 & 6.88                      & 13.74          & 14.71M & 2.55B & 13.68 \\
        \checkmark                   & \checkmark                    & \checkmark                  &                             & 13.71                     & 41.3                  & 7.03                      & 14.31          & 14.71M & 1.34B & 12.86 \\
        \checkmark                   & \checkmark                    & \checkmark                  & \checkmark                  & 14.27                     & 42.18                 & \textbf{6.79}             & \textbf{13.65} & 14.71M & 2.61B & 13.81 \\
        \midrule
        \multicolumn{8}{l}{\textit{RMSNorm Variants}}                                                                                                                                                                                                      \\
                                     &                               & \checkmark                  &                             & 13.69                     & 41.31                 & 6.85                      & 13.88          & 10.49M & 1.05B & 12.10 \\
                                     &                               & \checkmark                  & \checkmark                  & 13.82                     & 41.58                 & 6.94                      & 14.22          & 10.49M & 1.87B & 12.83 \\
        \checkmark                   &                               & \checkmark                  &                             & \textbf{12.95}            & \textbf{40.26}        & 6.80                      & 14.15          & 14.71M & 1.32B & 12.47 \\
        \checkmark                   &                               & \checkmark                  & \checkmark                  & 14.04                     & 41.99                 & 7.00                      & 13.96          & 14.71M & 2.55B & 13.62 \\
        \checkmark                   & \checkmark                    & \checkmark                  &                             & 13.20                     & 40.57                 & 6.91                      & 14.25          & 14.71M & 1.34B & 12.78 \\
        \checkmark                   & \checkmark                    & \checkmark                  & \checkmark                  & 13.71                     & 41.67                 & 6.86                      & 13.93          & 14.71M & 2.61B & 13.65 \\
        \bottomrule
    \end{tabular}
    \label{tab:architectural_variations}
\end{table}

The comparison between layer normalization and \ac{RMS} normalization highlights a key trade-off between feature stability and style robustness. The \ac{WD} split, which features known writers but unseen vocabulary, consistently favors variants with \ac{RMS} normalization. Because \ac{RMS} normalization scales activations without centering them, it preserves the natural variance of the text embeddings. By forcing the primary encoder to align with this highly expressive semantic space during training, the encoder learns to capture subtle, fine-grained character distinctions. This rich representation pays off during inference, allowing the model to flexibly compose familiar character patterns into entirely novel word sequences.

In contrast, the \ac{WI} split, which is heavily influenced by the noise of unseen handwriting styles, benefits from the stricter standardization of layer normalization. This method projects every text sample into a unified distribution, preventing the primary encoder from mapping global style shifts or writer-specific idiosyncrasies when aligning its features. Consequently, the encoder learns a robust, style-agnostic representation that generalizes well to unknown writers at deployment.

The addition of gated attention generally acts as a signal sharpener. Across most \ac{RMS} configurations, gated attention improves performance by dynamically suppressing background noise in the sensor and text embeddings before alignment. This is particularly effective in the \ac{WD} split, where the ``RMSNorm + Gated Attention'' combination yields the most significant gains. It forces the primary encoder to learn a precise, expressive representation that handles unseen word compositions at inference time.

The role of registers is highly context-dependent. On the \ac{WD} split, they provide no measurable benefit and often slightly degrade performance, suggesting that over-purifying the text embeddings might strip away useful global context when writer characteristics are already familiar. However, on the \ac{WI} split, the combination of ``LayerNorm + Gated Attention + Registers'' achieves the best overall results. This suggests that register tokens act as ``sinks'' for global textual artifacts, ensuring the resulting semantic text tokens are exceptionally pure and localized. When the primary encoder is forced to align with this strictly purified target space during training, it is discouraged from relying on spurious global correlations, such as an individual's overarching writing style.

While these architectural enhancements drive substantial accuracy improvements, they also reveal a notable dynamic regarding training-time computational overhead. When integrating the fully equipped auxiliary branch, the parameter count increases by roughly 274\% and the \ac{MACs} surge by over 330\% compared to the baseline. Yet, the actual training latency only increases by 60\%. This disparity between the explosion of \ac{MACs} and the modest bump in latency stems from the architecture itself. The base model relies on a bidirectional \ac{LSTM} decoder, which requires sequential processing and creates a strict temporal bottleneck on the hardware. Conversely, the Transformer and attention components of the auxiliary branch are highly parallelizable. Thus, the GPU can readily absorb the massive computational bulk of the auxiliary branch without suffering a proportional hit to training speed.

\subsection{Ablation Study on Error Set Size}

\begin{figure}[h]
    \centering
    \begin{tikzpicture}
        \begin{groupplot}[
                group style={
                        group size=2 by 2,
                        horizontal sep=2cm,
                        vertical sep=1cm
                    },
                width=0.46\linewidth,
                height=4cm,
                xtick={1, 2, 3},
                xticklabels={1, 2, 3},
                ymajorgrids=true,
                grid style={dashed, gray!30},
            ]

            \nextgroupplot[
                title={\textbf{\ac{CER} on the \ac{WD} Split}},
                ylabel={Error Rates (\%)},
                xticklabels={},
                legend columns=1,
                legend style={
                        at={(1.248, 0.5)},
                        anchor=center,
                        fill=none,
                        draw=none,
                        nodes={scale=0.5, transform shape}
                    }
            ]
            \addplot[color=blue, mark=square*, thick, opacity=0.5, mark options={opacity=0.5}]
            coordinates {(1, 14.00)(2, 13.81)(3, 13.85)};
            \addlegendentry{LN}
            \addplot[color=red, mark=triangle*, thick, opacity=0.5, mark options={opacity=0.5}]
            coordinates {(1, 14.30)(2, 13.82)(3, 14.15)};
            \addlegendentry{RMS}
            \addplot[color=green!60!black, mark=*, thick, opacity=0.5, mark options={opacity=0.5}]
            coordinates {(1, 14.20)(2, 13.94)(3, 14.55)};
            \addlegendentry{LN+GA}
            \addplot[color=orange, mark=diamond*, thick, opacity=0.5, mark options={opacity=0.5}]
            coordinates {(1, 13.71)(2, 14.04)(3, 14.14)};
            \addlegendentry{RMS+GA}
            \addplot[color=violet, mark=pentagon*, thick, opacity=0.5, mark options={opacity=0.5}]
            coordinates {(1, 13.79)(2, 14.27)(3, 14.38)};
            \addlegendentry{LN+GA+Reg}
            \addplot[color=teal, mark=star, thick, opacity=0.5, mark options={opacity=0.5}]
            coordinates {(1, 14.21)(2, 13.71)(3, 14.12)};
            \addlegendentry{RMS+GA+Reg}

            \nextgroupplot[
                title={\textbf{\ac{WER} on the \ac{WD} Split}},
                xticklabels={},
                yticklabel pos=right
            ]
            \addplot[color=blue, mark=square*, thick, opacity=0.5, mark options={opacity=0.5}]
            coordinates {(1, 41.98)(2, 41.74)(3, 41.24)};
            \addplot[color=red, mark=triangle*, thick, opacity=0.5, mark options={opacity=0.5}]
            coordinates {(1, 42.93)(2, 41.58)(3, 42.29)};
            \addplot[color=green!60!black, mark=*, thick, opacity=0.5, mark options={opacity=0.5}]
            coordinates {(1, 42.80)(2, 41.82)(3, 42.93)};
            \addplot[color=orange, mark=diamond*, thick, opacity=0.5, mark options={opacity=0.5}]
            coordinates {(1, 41.55)(2, 41.99)(3, 42.23)};
            \addplot[color=violet, mark=pentagon*, thick, opacity=0.5, mark options={opacity=0.5}]
            coordinates {(1, 41.74)(2, 42.18)(3, 41.89)};
            \addplot[color=teal, mark=star, thick, opacity=0.5, mark options={opacity=0.5}]
            coordinates {(1, 42.27)(2, 41.67)(3, 41.99)};

            \nextgroupplot[
                title={\textbf{\ac{CER} on the \ac{WI} Split}},
                xlabel={\#Error Sets},
                ylabel={Error Rates (\%)},
                legend columns=1,
                legend style={
                        at={(1.248, 0.5)},
                        anchor=center,
                        fill=none,
                        draw=none,
                        nodes={scale=0.5, transform shape}
                    }
            ]
            \addplot[color=blue, mark=square*, thick, opacity=0.5, mark options={opacity=0.5}]
            coordinates {(1, 7.04)(2, 7.10)(3, 6.85)};
            \addlegendentry{LN}
            \addplot[color=red, mark=triangle*, thick, opacity=0.5, mark options={opacity=0.5}]
            coordinates {(1, 6.96)(2, 6.94)(3, 7.04)};
            \addlegendentry{RMS}
            \addplot[color=green!60!black, mark=*, thick, opacity=0.5, mark options={opacity=0.5}]
            coordinates {(1, 6.81)(2, 6.88)(3, 6.93)};
            \addlegendentry{LN+GA}
            \addplot[color=orange, mark=diamond*, thick, opacity=0.5, mark options={opacity=0.5}]
            coordinates {(1, 6.99)(2, 7.00)(3, 7.05)};
            \addlegendentry{RMS+GA}
            \addplot[color=violet, mark=pentagon*, thick, opacity=0.5, mark options={opacity=0.5}]
            coordinates {(1, 6.98)(2, 6.79)(3, 7.02)};
            \addlegendentry{LN+GA+Reg}
            \addplot[color=teal, mark=star, thick, opacity=0.5, mark options={opacity=0.5}]
            coordinates {(1, 6.92)(2, 6.86)(3, 6.93)};
            \addlegendentry{RMS+GA+Reg}

            \nextgroupplot[
                title={\textbf{\ac{WER} on the \ac{WI} Split}},
                xlabel={\#Error Sets},
                yticklabel pos=right
            ]
            \addplot[color=blue, mark=square*, thick, opacity=0.5, mark options={opacity=0.5}]
            coordinates {(1, 14.06)(2, 14.31)(3, 13.90)};
            \addplot[color=red, mark=triangle*, thick, opacity=0.5, mark options={opacity=0.5}]
            coordinates {(1, 14.26)(2, 14.22)(3, 14.07)};
            \addplot[color=green!60!black, mark=*, thick, opacity=0.5, mark options={opacity=0.5}]
            coordinates {(1, 13.91)(2, 13.74)(3, 13.99)};
            \addplot[color=orange, mark=diamond*, thick, opacity=0.5, mark options={opacity=0.5}]
            coordinates {(1, 14.20)(2, 13.96)(3, 14.13)};
            \addplot[color=violet, mark=pentagon*, thick, opacity=0.5, mark options={opacity=0.5}]
            coordinates {(1, 13.65)(2, 13.65)(3, 13.85)};
            \addplot[color=teal, mark=star, thick, opacity=0.5, mark options={opacity=0.5}]
            coordinates {(1, 13.84)(2, 13.93)(3, 13.81)};

        \end{groupplot}
    \end{tikzpicture}
    \caption{\textbf{Sensitivity to Negative Sample Diversity.} The plots show performance changes across different error set sizes. The central legend applies to all subplots.}
    \label{fig:error_set_size}
\end{figure}
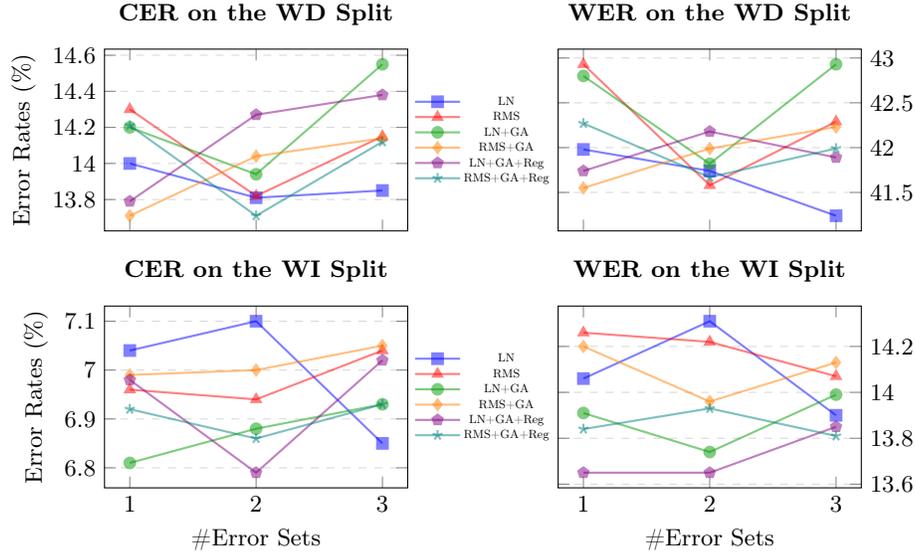

To determine the optimal task difficulty for the error-based contrastive objective, we investigate how the framework responds to the number of hard negative samples. This analysis is conducted across various \ac{ECHWR} configurations, with the results shown in Fig.~\ref{fig:error_set_size}.

Across most architectural configurations, an error set size of two emerges as the most effective setting. On both the \ac{WD} and \ac{WI} splits, we observe a general trend where performance improves significantly when increasing the number of error sets from one to two. However, accuracy often plateaus or declines when increasing the count to three.

The results reveal an interesting relationship between architectural complexity and the need for negative samples. Simpler models often show continued improvement or stability with three error sets. Because they lack sophisticated attention filtering, these models seem to benefit from the stronger data augmentation provided by a larger quantity of hard negatives. In contrast, models equipped with gated attention or registers tend to peak at two error sets. Since these mechanisms are already efficient at suppressing noise and focusing on informative features, adding more negative samples yields diminishing returns. Therefore, we use an error set size of two as the standard configuration for \ac{ECHWR}.

\subsection{Qualitative Analysis}

\begin{figure}[ht]
    \centering
    \includegraphics[width=0.45\linewidth]{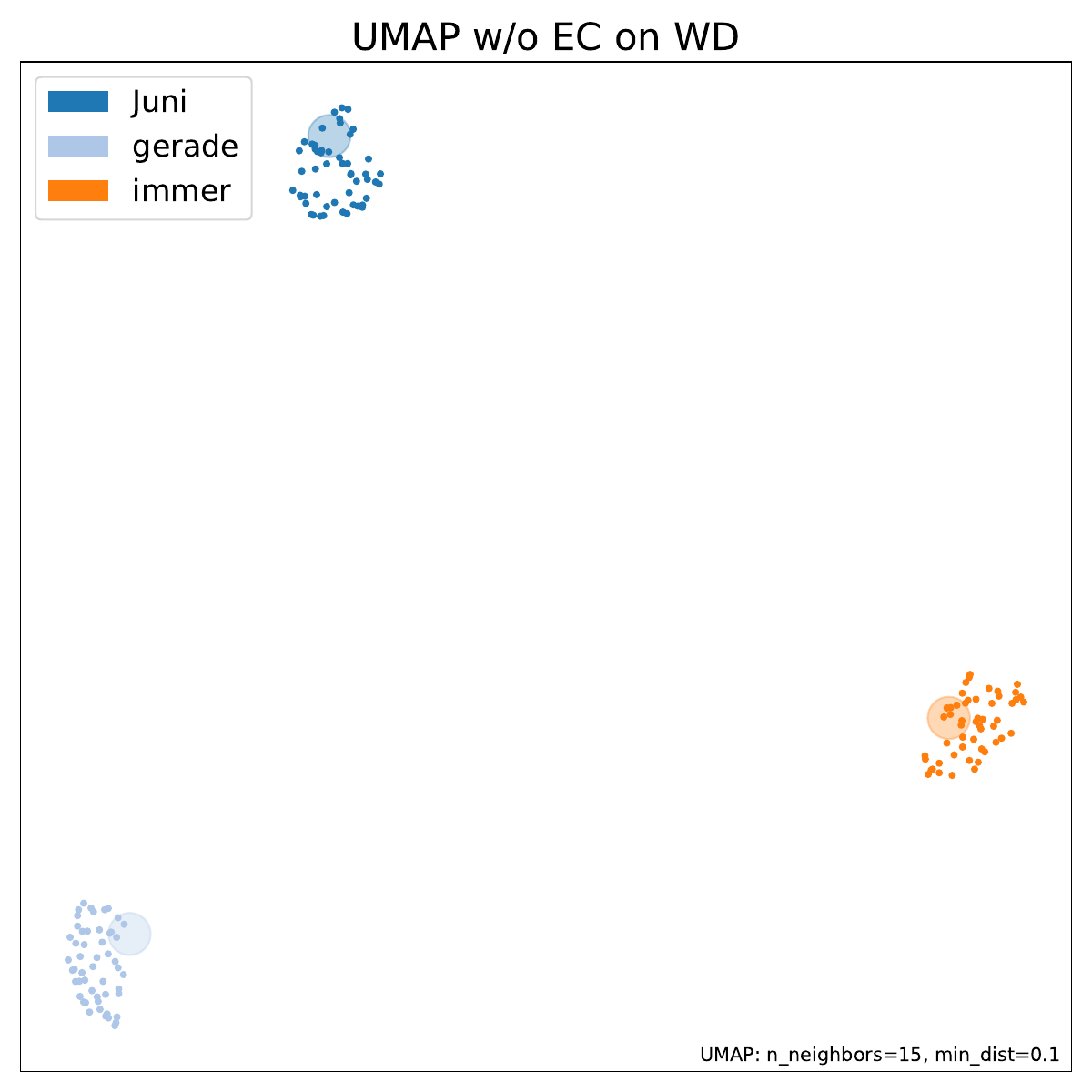}
    \includegraphics[width=0.45\linewidth]{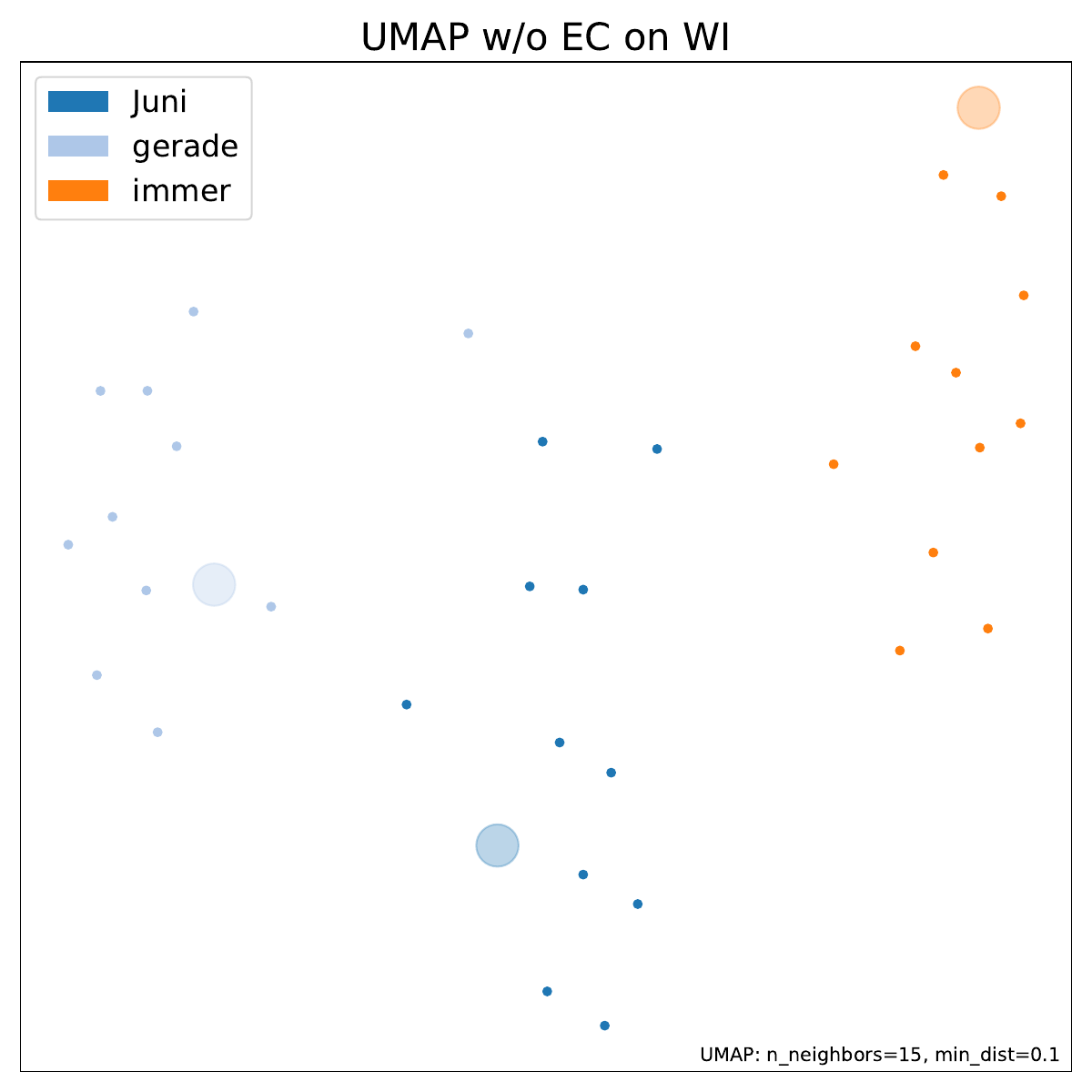}
    \includegraphics[width=0.45\linewidth]{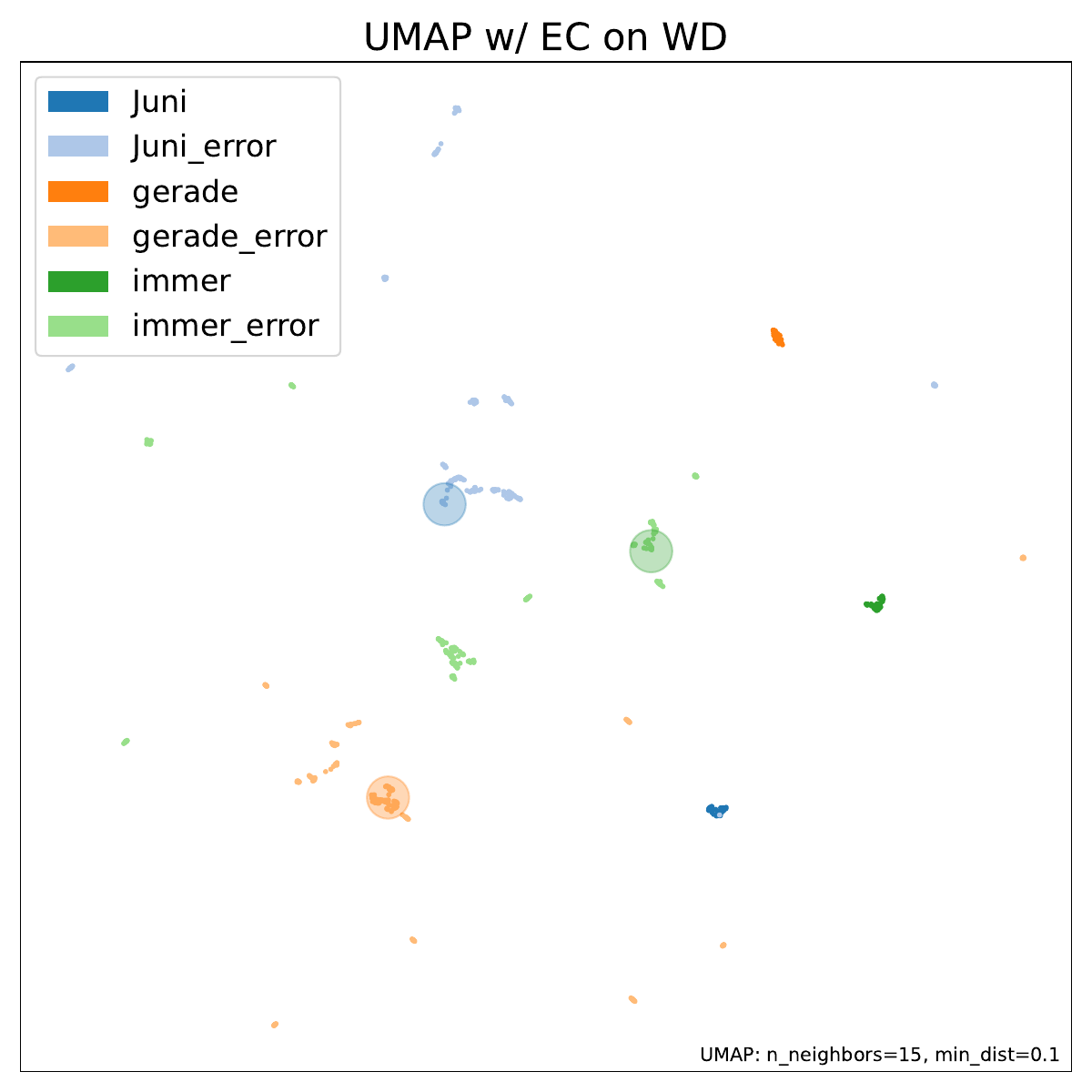}
    \includegraphics[width=0.45\linewidth]{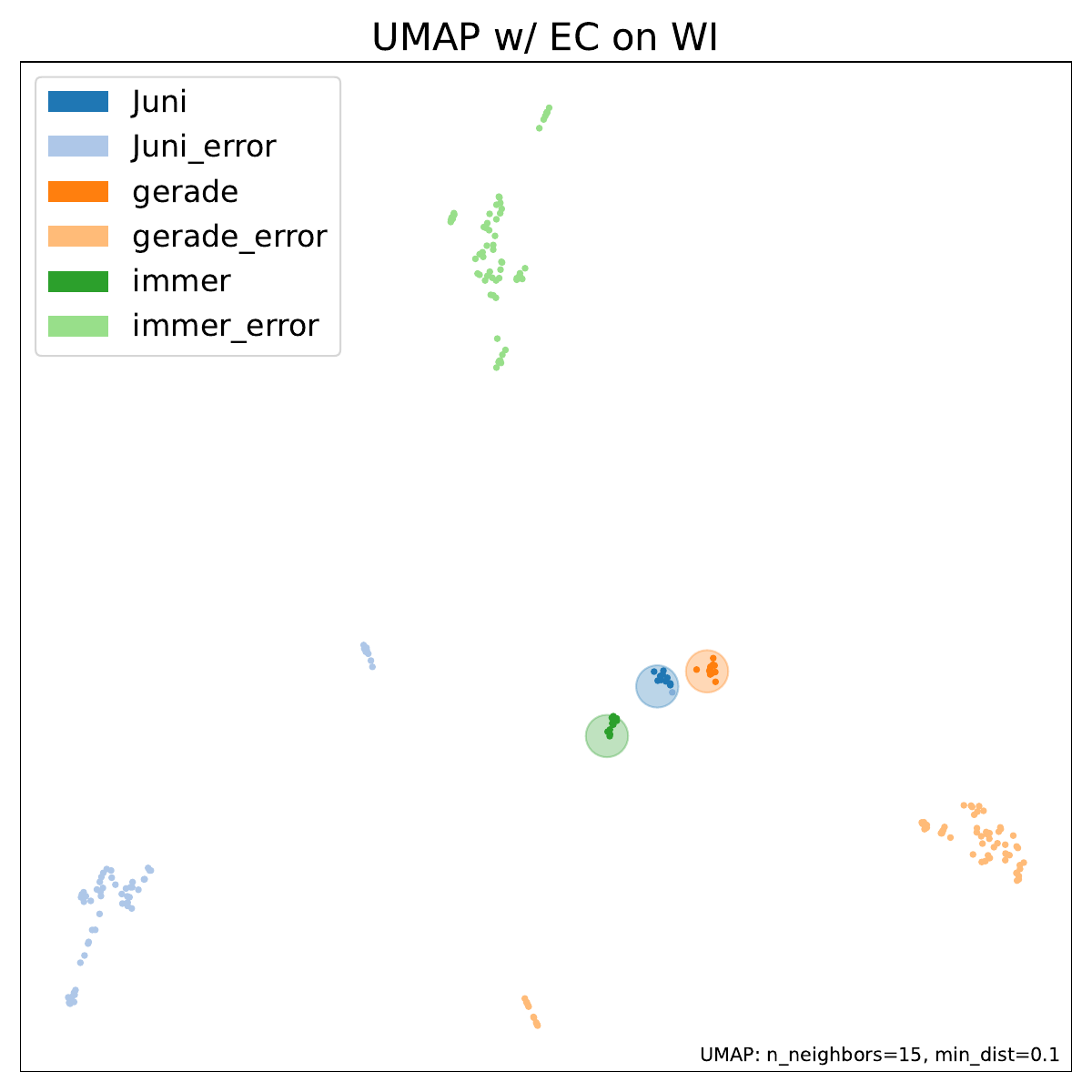}
    \caption{\textbf{UMAP Visualization of the Embedding Space.} The plots display projected embeddings of sensor sequences (small dots), ground-truth text anchors (large transparent circles), and hard negative errors (faded dots). The top row shows baseline embeddings from models trained with the \ac{BC} objective only, while the bottom row illustrates the impact of adding the \ac{EC} objective.}
    \label{fig:umap}
\end{figure}

To empirically validate the impact of the contrastive objectives, we visualize the learned embedding space using UMAP~\cite{umap}. We extract embeddings from the validation sets of the first fold using the optimal configurations: \textit{RMSNorm + Gated Attention} for the \ac{WD} split and \textit{LayerNorm + Gated Attention + Registers} for the \ac{WI} split. Figure~\ref{fig:umap} presents the projections of sensor sequences, ground-truth text anchors, and generated hard negative errors for three sample words.

The visualization highlights a distinct structural divergence between the two generalization tasks. In the baseline configurations without \ac{EC}, the sensor embeddings generally form cohesive clusters that align with their corresponding ground-truth text anchors. This indicates that the \ac{CTC} and \ac{BC} objectives provide effective modality alignment.

The impact of the \ac{EC} objective is most evident in the second row. For the \ac{WI} split, where the vocabulary is known but the writers are new, the \ac{EC} objective successfully enforces a safety margin. The embeddings of hard negatives are pushed significantly away from the true positive clusters. This explicit separation reduces ambiguity and ensures that valid handwriting trajectories are not confused with fine-grained typos, which explains the superior performance of \ac{EC} on the \ac{WI} split.

In contrast, the \ac{WD} split reveals why the \ac{EC} objective is less effective for unseen vocabulary. Here, we observe a significant misalignment between the text anchors and their corresponding sensor sequence clusters. This discrepancy arises because the text encoder, having been trained on a specific character distribution, struggles to generalize its embeddings to the unseen word combinations in the validation set. This confirms our hypothesis that the flexible alignment of \ac{BC} is preferred for the \ac{WD} split, whereas the rigid discrimination of \ac{EC} requires the stable anchors found in known vocabulary.

\section{Conclusion \& Outlook}

In this work, we presented \ac{ECHWR}, a novel training framework designed to enhance \ac{IMU}-based online handwriting recognition. By introducing a temporary auxiliary branch and a dual contrastive objective, we enable standard encoder-decoder models to learn richer, more discriminative representations. Crucially, this improvement is achieved without modifying the baseline \ac{OnHWR} model or incurring additional computational costs during inference. Our experiments on the OnHW-Words500 dataset demonstrate that \ac{ECHWR} significantly outperforms state-of-the-art baselines while maintaining the exact same inference latency as the base model.

Our ablation studies reveal a critical dichotomy between different generalization tasks. We found that the \ac{WD} task, which requires generalizing to unseen vocabulary, benefits most from the feature stability provided by \ac{RMS} normalization and \ac{BC} alignment. In contrast, the \ac{WI} task is dominated by the variance of unseen writers. This scenario demands the robustness of layer normalization with registers and the fine-grained discrimination enforced by the \ac{EC} objective.

Looking forward, we acknowledge that this study is limited by current dataset availability. This constraint prevented the identification of a universal configuration capable of simultaneously handling both writing style variations and character distribution shifts. Future work should address this limitation by leveraging larger-scale datasets with more complex vocabularies to determine whether a unified architecture can bridge this gap.

\clearpage
\appendix
\section{Visualization of Character Distribution}
\label{app:visualize_character_distribution}

\begin{figure}[ht]
    \centering
    \includegraphics[width=\linewidth]{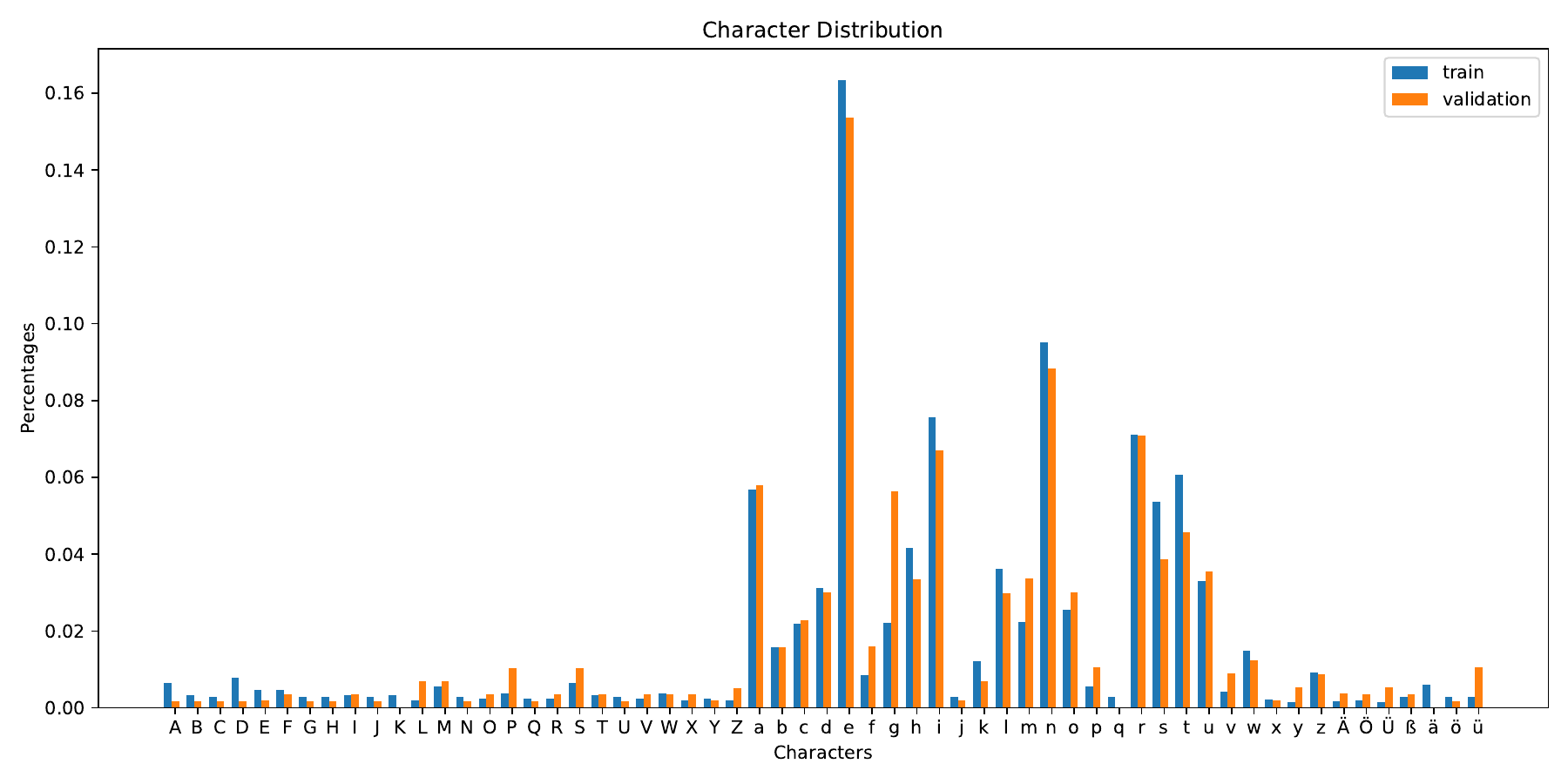}
    \includegraphics[width=\linewidth]{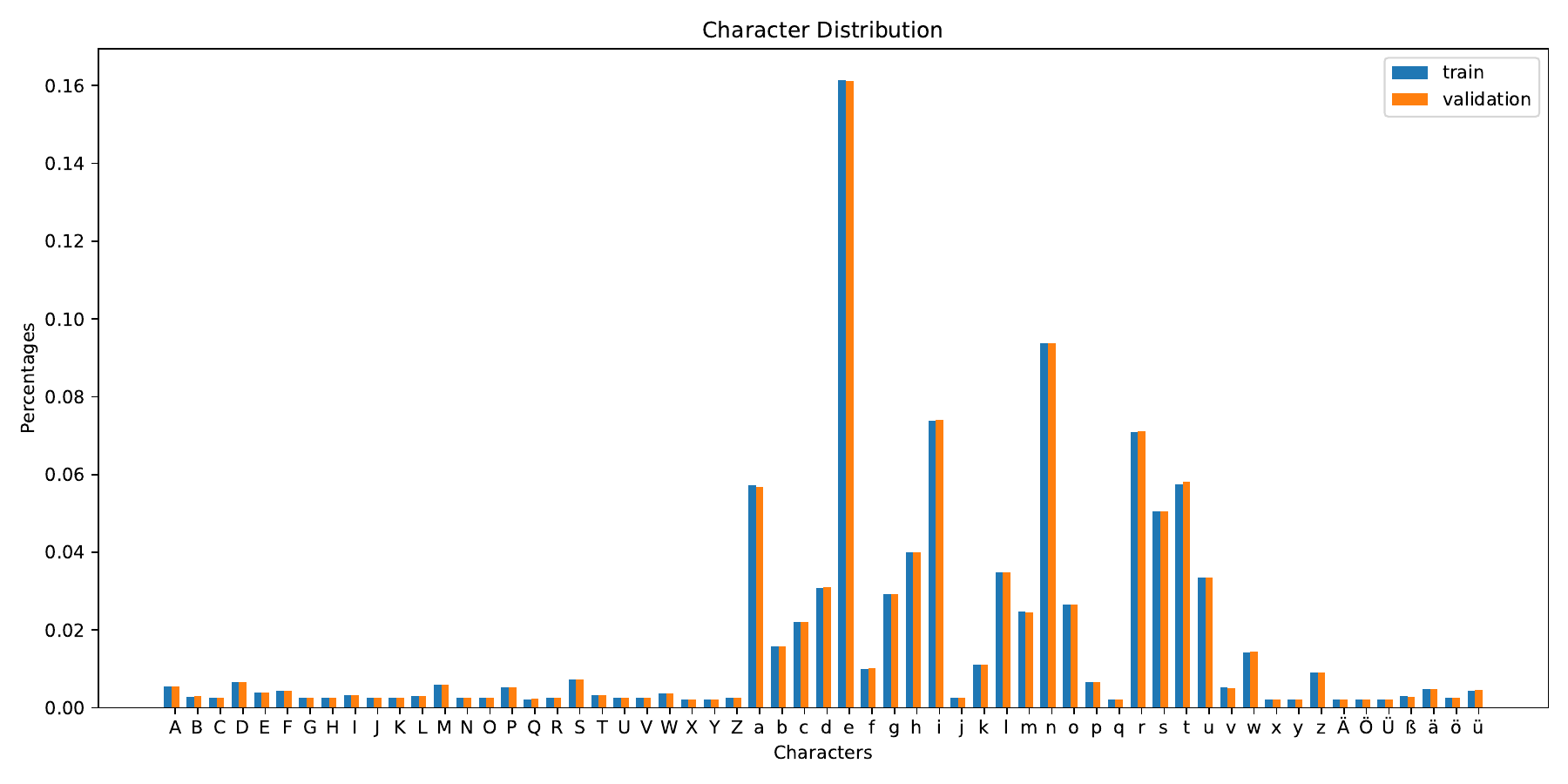}
    \caption{\textbf{Character distribution of the right-handed OnHW Words500 dataset.} The upper and lower plots show the character distributions for the first fold of the \ac{WD} and \ac{WI} splits, respectively. Blue bars represent character frequencies in the training sets, while orange bars represent frequencies in the validation sets.}
    \label{fig:freq_char}
\end{figure}

As demonstrated in Fig.~\ref{fig:freq_char}, the character distributions of the \ac{WD} and \ac{WI} splits in the OnHWR-Words500 dataset differ significantly.

In the \ac{WI} split (lower), the distributions between the training and validation sets match almost perfectly. Furthermore, each character is represented by at least a few data samples in both sets. In contrast, the distributions for the \ac{WD} split (upper) vary greatly between the training and validation sets. Certain characters, such as ``q'' and ``ä'', do not appear in the validation set at all.

\newpage
%
%
\bibliographystyle{splncs04}
\bibliography{references}

@inbook{privacy1,
  author    = {Marcos Faundez-Zanuy and Jiri Mekyska},
  title     = {Privacy of online handwriting biometrics related to biomedical analysis},
  booktitle = {User-Centric Privacy and Security in Biometrics},
  chapter   = {Chapter 2},
  pages     = {17-39},
  year      = {2017},
  publisher = {IET},
  doi       = {10.1049/PBSE004E_ch2},
  url       = {https://digital-library.theiet.org/doi/abs/10.1049/PBSE004E_ch2}
}

@article{privacy2,
  author  = {Zhang, Peirong and Liu, Yuliang and Lai, Songxuan and Li, Hongliang and Jin, Lianwen},
  journal = {IEEE Transactions on Pattern Analysis and Machine Intelligence},
  title   = {Privacy-Preserving Biometric Verification With Handwritten Random Digit String},
  year    = {2025},
  volume  = {47},
  number  = {4},
  pages   = {3049-3066},
  doi     = {10.1109/TPAMI.2025.3529022}
}

@inproceedings{origaminet,
  author    = {Yousef, Mohamed and Bishop, Tom E.},
  booktitle = {2020 IEEE/CVF Conference on Computer Vision and Pattern Recognition (CVPR)},
  title     = {OrigamiNet: Weakly-Supervised, Segmentation-Free, One-Step, Full Page Text Recognition by learning to unfold},
  year      = {2020},
  volume    = {},
  number    = {},
  pages     = {14698-14707},
  doi       = {10.1109/CVPR42600.2020.01472}
}

@inproceedings{trocr,
  author    = {Li, Minghao and Lv, Tengchao and Chen, Jingye and Cui, Lei and Lu, Yijuan and Florencio, Dinei and Zhang, Cha and Li, Zhoujun and Wei, Furu},
  title     = {TrOCR: transformer-based optical character recognition with pre-trained models},
  year      = {2023},
  isbn      = {978-1-57735-880-0},
  publisher = {AAAI Press},
  url       = {https://doi.org/10.1609/aaai.v37i11.26538},
  doi       = {10.1609/aaai.v37i11.26538},
  booktitle = {Proceedings of the Thirty-Seventh AAAI Conference on Artificial Intelligence and Thirty-Fifth Conference on Innovative Applications of Artificial Intelligence and Thirteenth Symposium on Educational Advances in Artificial Intelligence},
  articleno = {1469},
  numpages  = {9},
  series    = {AAAI'23/IAAI'23/EAAI'23}
}

@article{htr-vt,
  title   = {HTR-VT: Handwritten text recognition with vision transformer},
  journal = {Pattern Recognition},
  volume  = {158},
  pages   = {110967},
  year    = {2025},
  issn    = {0031-3203},
  doi     = {https://doi.org/10.1016/j.patcog.2024.110967},
  url     = {https://www.sciencedirect.com/science/article/pii/S0031320324007180},
  author  = {Yuting Li and Dexiong Chen and Tinglong Tang and Xi Shen}
}

@inproceedings{simclr,
  author    = {Chen, Ting and Kornblith, Simon and Norouzi, Mohammad and Hinton, Geoffrey},
  title     = {A simple framework for contrastive learning of visual representations},
  year      = {2020},
  publisher = {JMLR.org},
  booktitle = {Proceedings of the 37th International Conference on Machine Learning},
  articleno = {149},
  numpages  = {11},
  series    = {ICML'20}
}

@inproceedings{clip,
  title     = {Learning Transferable Visual Models From Natural Language Supervision},
  author    = {Radford, Alec and Kim, Jong Wook and Hallacy, Chris and Ramesh, Aditya and Goh, Gabriel and Agarwal, Sandhini and Sastry, Girish and Askell, Amanda and Mishkin, Pamela and Clark, Jack and Krueger, Gretchen and Sutskever, Ilya},
  booktitle = {Proceedings of the 38th International Conference on Machine Learning},
  pages     = {8748--8763},
  year      = {2021},
  editor    = {Meila, Marina and Zhang, Tong},
  volume    = {139},
  series    = {Proceedings of Machine Learning Research},
  month     = {18--24 Jul},
  publisher = {PMLR},
  url       = {https://proceedings.mlr.press/v139/radford21a.html}
}

@inproceedings{cl_pt_1,
  title     = {Time-Series Representation Learning via Temporal and Contextual Contrasting},
  author    = {Eldele, Emadeldeen and Ragab, Mohamed and Chen, Zhenghua and Wu, Min and Kwoh, Chee Keong and Li, Xiaoli and Guan, Cuntai},
  booktitle = {Proceedings of the Thirtieth International Joint Conference on Artificial Intelligence, {IJCAI-21}},
  pages     = {2352--2359},
  year      = {2021}
}

@inproceedings{imu_svm_hmm,
  author    = {Amma, Christoph and Georgi, Marcus and Schultz, Tanja},
  booktitle = {2012 16th International Symposium on Wearable Computers},
  title     = {Airwriting: Hands-Free Mobile Text Input by Spotting and Continuous Recognition of 3d-Space Handwriting with Inertial Sensors},
  year      = {2012},
  volume    = {},
  number    = {},
  pages     = {52-59},
  doi       = {10.1109/ISWC.2012.21}
}

@inproceedings{imu_hmm_dtw,
  author    = {Choi, Sung-do and Lee, Alexander S. and Lee, Soo-young},
  booktitle = {2006 IEEE International Conference on Information Acquisition},
  title     = {On-Line Handwritten Character Recognition with 3D Accelerometer},
  year      = {2006},
  volume    = {},
  number    = {},
  pages     = {845-850},
  doi       = {10.1109/ICIA.2006.305842}
}

@article{imu_dtw_1,
  author  = {Hsu, Yu-Liang and Chu, Cheng-Ling and Tsai, Yi-Ju and Wang, Jeen-Shing},
  journal = {IEEE Sensors Journal},
  title   = {An Inertial Pen With Dynamic Time Warping Recognizer for Handwriting and Gesture Recognition},
  year    = {2015},
  volume  = {15},
  number  = {1},
  pages   = {154-163},
  doi     = {10.1109/JSEN.2014.2339843}
}

@inproceedings{imu_dtw_2,
  title     = {Online Handwriting Recognition Using an Accelerometer-Based Pen Device},
  author    = {Wang Jeen-Shing and Hsu Yu-Liang and Chu Cheng-Ling},
  year      = {2013},
  booktitle = {Proceedings of the 2nd International Conference on Advances in Computer Science and Engineering (CSE 2013)},
  pages     = {231-234},
  issn      = {1951-6851},
  isbn      = {978-90786-77-70-3},
  url       = {https://doi.org/10.2991/cse.2013.52},
  doi       = {10.2991/cse.2013.52},
  publisher = {Atlantis Press}
}

@inproceedings{imu_cnn_lstm_1,
  author    = {Wehbi, Mohamad and Hamann, Tim and Barth, Jens and Kaempf, Peter and Zanca, Dario and Eskofier, Bjoern},
  editor    = {Llad{\'o}s, Josep and Lopresti, Daniel and Uchida, Seiichi},
  title     = {Towards an IMU-based Pen Online Handwriting Recognizer},
  booktitle = {Document Analysis and Recognition -- ICDAR 2021},
  year      = {2021},
  publisher = {Springer International Publishing},
  address   = {Cham},
  pages     = {289--303},
  isbn      = {978-3-030-86334-0},
  doi       = {10.1007/978-3-030-86334-0_19}
}

@article{imu_cnn_lstm_2,
  author     = {Ott, Felix and R\"{u}gamer, David and Heublein, Lucas and Hamann, Tim and Barth, Jens and Bischl, Bernd and Mutschler, Christopher},
  title      = {Benchmarking online sequence-to-sequence and character-based handwriting recognition from IMU-enhanced pens},
  year       = {2022},
  issue_date = {Dec 2022},
  publisher  = {Springer-Verlag},
  address    = {Berlin, Heidelberg},
  volume     = {25},
  number     = {4},
  issn       = {1433-2833},
  url        = {https://doi.org/10.1007/s10032-022-00415-6},
  doi        = {10.1007/s10032-022-00415-6},
  journal    = {Int. J. Doc. Anal. Recognit.},
  month      = dec,
  pages      = {385-414},
  numpages   = {30}
}

@inproceedings{rewi,
  author    = {Li, Jindong
               and Hamann, Tim
               and Barth, Jens
               and K{\"a}mpf, Peter
               and Zanca, Dario
               and Eskofier, Bj{\"o}rn},
  editor    = {Durmaz Incel, {\"O}zlem
               and Qin, Jingwen
               and Bieber, Gerald
               and Kuijper, Arjan},
  title     = {Robust and Efficient Writer-Independent IMU-Based Handwriting Recognition},
  booktitle = {Sensor-Based Activity Recognition and Artificial Intelligence},
  year      = {2026},
  publisher = {Springer Nature Switzerland},
  address   = {Cham},
  pages     = {261--286},
  isbn      = {978-3-032-13312-0}
}

@inproceedings{cl_pt_2,
  author    = {Zhang, Xiang and Zhao, Ziyuan and Tsiligkaridis, Theodoros and Zitnik, Marinka},
  title     = {Self-supervised contrastive pre-training for time series via time-frequency consistency},
  year      = {2022},
  isbn      = {9781713871088},
  publisher = {Curran Associates Inc.},
  address   = {Red Hook, NY, USA},
  booktitle = {Proceedings of the 36th International Conference on Neural Information Processing Systems},
  articleno = {288},
  numpages  = {16},
  location  = {New Orleans, LA, USA},
  series    = {NIPS '22}
}

@inproceedings{cl_pt_3,
  title        = {Clocs: Contrastive learning of cardiac signals across space, time, and patients},
  author       = {Kiyasseh, Dani and Zhu, Tingting and Clifton, David A},
  booktitle    = {International Conference on Machine Learning},
  pages        = {5606--5615},
  year         = {2021},
  organization = {PMLR}
}

@inproceedings{cl_pt_4,
  author    = {Khaertdinov, Bulat and Asteriadis, Stylianos},
  booktitle = {2022 IEEE International Joint Conference on Biometrics (IJCB)},
  title     = {Temporal Feature Alignment in Contrastive Self-Supervised Learning for Human Activity Recognition},
  year      = {2022},
  volume    = {},
  number    = {},
  pages     = {1-9},
  doi       = {10.1109/IJCB54206.2022.10007984}
}

@inproceedings{cl_mt_1,
  author    = {Liu, Xu and Liang, Yuxuan and Huang, Chao and Zheng, Yu and Hooi, Bryan and Zimmermann, Roger},
  title     = {When do contrastive learning signals help spatio-temporal graph forecasting?},
  year      = {2022},
  isbn      = {9781450395298},
  publisher = {Association for Computing Machinery},
  address   = {New York, NY, USA},
  url       = {https://doi.org/10.1145/3557915.3560939},
  doi       = {10.1145/3557915.3560939},
  booktitle = {Proceedings of the 30th International Conference on Advances in Geographic Information Systems},
  articleno = {5},
  numpages  = {12},
  location  = {Seattle, Washington},
  series    = {SIGSPATIAL '22}
}

@article{cl_mt_2,
  author         = {Guo, Pengyu and Nakayama, Masaya},
  title          = {Towards User-Generalizable Wearable-Sensor-Based Human Activity Recognition: A Multi-Task Contrastive Learning Approach},
  journal        = {Sensors},
  volume         = {25},
  year           = {2025},
  number         = {22},
  article-number = {6988},
  url            = {https://www.mdpi.com/1424-8220/25/22/6988},
  pubmedid       = {41305195},
  issn           = {1424-8220},
  doi            = {10.3390/s25226988}
}

@inproceedings{hard_neg_1,
  author    = {Jiang, Ruijie and Nguyen, Thuan and Ishwar, Prakash and Aeron, Shuchin},
  booktitle = {2024 International Joint Conference on Neural Networks (IJCNN)},
  title     = {Supervised Contrastive Learning with Hard Negative Samples},
  year      = {2024},
  pages     = {1-8},
  doi       = {10.1109/IJCNN60899.2024.10650863}
}

@inproceedings{hard_neg_2,
  author    = {Kalantidis, Yannis and Sariyildiz, Mert Bulent and Pion, Noe and Weinzaepfel, Philippe and Larlus, Diane},
  title     = {Hard negative mixing for contrastive learning},
  year      = {2020},
  isbn      = {9781713829546},
  publisher = {Curran Associates Inc.},
  address   = {Red Hook, NY, USA},
  booktitle = {Proceedings of the 34th International Conference on Neural Information Processing Systems},
  articleno = {1829},
  numpages  = {12},
  location  = {Vancouver, BC, Canada},
  series    = {NIPS '20}
}

@inproceedings{hard_neg_3,
  title     = {Contrastive Learning with Hard Negative Samples},
  author    = {Joshua David Robinson and Ching-Yao Chuang and Suvrit Sra and Stefanie Jegelka},
  booktitle = {International Conference on Learning Representations},
  year      = {2021},
  url       = {https://openreview.net/forum?id=CR1XOQ0UTh-}
}

@inproceedings{trans,
  author    = {Vaswani, Ashish and Shazeer, Noam and Parmar, Niki and Uszkoreit, Jakob and Jones, Llion and Gomez, Aidan N and Kaiser, \L ukasz and Polosukhin, Illia},
  booktitle = {Advances in Neural Information Processing Systems},
  editor    = {I. Guyon and U. Von Luxburg and S. Bengio and H. Wallach and R. Fergus and S. Vishwanathan and R. Garnett},
  pages     = {},
  publisher = {Curran Associates, Inc.},
  title     = {Attention is All you Need},
  url       = {https://proceedings.neurips.cc/paper_files/paper/2017/file/3f5ee243547dee91fbd053c1c4a845aa-Paper.pdf},
  volume    = {30},
  year      = {2017}
}

@inproceedings{gated,
  title     = {Gated Attention for Large Language Models: Non-linearity, Sparsity, and Attention-Sink-Free},
  author    = {Zihan Qiu and Zekun Wang and Bo Zheng and Zeyu Huang and Kaiyue Wen and Songlin Yang and Rui Men and Le Yu and Fei Huang and Suozhi Huang and Dayiheng Liu and Jingren Zhou and Junyang Lin},
  booktitle = {The Thirty-ninth Annual Conference on Neural Information Processing Systems},
  year      = {2025},
  url       = {https://openreview.net/forum?id=1b7whO4SfY}
}

@inproceedings{register,
  title     = {Vision Transformers Need Registers},
  author    = {Timoth{\'e}e Darcet and Maxime Oquab and Julien Mairal and Piotr Bojanowski},
  booktitle = {The Twelfth International Conference on Learning Representations},
  year      = {2024},
  url       = {https://openreview.net/forum?id=2dnO3LLiJ1}
}

@inproceedings{rmsnorm,
  author    = {Zhang, Biao and Sennrich, Rico},
  booktitle = {Advances in Neural Information Processing Systems},
  editor    = {H. Wallach and H. Larochelle and A. Beygelzimer and F. d\textquotesingle Alch\'{e}-Buc and E. Fox and R. Garnett},
  pages     = {},
  publisher = {Curran Associates, Inc.},
  title     = {Root Mean Square Layer Normalization},
  url       = {https://proceedings.neurips.cc/paper_files/paper/2019/file/1e8a19426224ca89e83cef47f1e7f53b-Paper.pdf},
  volume    = {32},
  year      = {2019}
}

@inproceedings{convnext,
  author    = {Liu, Zhuang and Mao, Hanzi and Wu, Chao-Yuan and Feichtenhofer, Christoph and Darrell, Trevor and Xie, Saining},
  booktitle = {2022 IEEE/CVF Conference on Computer Vision and Pattern Recognition (CVPR)},
  title     = {A ConvNet for the 2020s},
  year      = {2022},
  volume    = {},
  number    = {},
  pages     = {11966-11976},
  doi       = {10.1109/CVPR52688.2022.01167}
}

@inproceedings{swinv2,
  author    = {Liu, Ze and Hu, Han and Lin, Yutong and Yao, Zhuliang and Xie, Zhenda and Wei, Yixuan and Ning, Jia and Cao, Yue and Zhang, Zheng and Dong, Li and Wei, Furu and Guo, Baining},
  booktitle = {2022 IEEE/CVF Conference on Computer Vision and Pattern Recognition (CVPR)},
  title     = {Swin Transformer V2: Scaling Up Capacity and Resolution},
  year      = {2022},
  volume    = {},
  number    = {},
  pages     = {11999-12009},
  doi       = {10.1109/CVPR52688.2022.01170}
}

@article{umap,
  doi       = {10.21105/joss.00861},
  url       = {https://doi.org/10.21105/joss.00861},
  year      = {2018},
  publisher = {The Open Journal},
  volume    = {3},
  number    = {29},
  pages     = {861},
  author    = {McInnes, Leland and Healy, John and Saul, Nathaniel and Großberger, Lukas},
  title     = {UMAP: Uniform Manifold Approximation and Projection},
  journal   = {Journal of Open Source Software}
}

\end{document}